\documentclass[lettersize,journal]{IEEEtran}

\usepackage{amsmath,amsfonts,amssymb,mathtools}
\usepackage{graphicx}
\usepackage{booktabs}
\usepackage{threeparttable}
\usepackage{array}
\usepackage{tabularx}
\usepackage{multirow}
\usepackage{makecell}
\usepackage{enumitem}
\usepackage[ruled,vlined,algo2e]{algorithm2e}
\usepackage{textcomp}
\usepackage{stfloats}
\usepackage{url}
\usepackage{cite}
\usepackage{siunitx}

\setlist[enumerate]{topsep=2pt,itemsep=1pt,parsep=0pt,partopsep=0pt}

\newcolumntype{C}[1]{>{\centering\arraybackslash}p{#1}}
\newcolumntype{L}[1]{>{\raggedright\arraybackslash}p{#1}}

\begin{document}

\title{Coverage-Aware Virtual IMU Augmentation for Low-Resource Human Activity Recognition}

\author{Jiayuan~Gao, Yingwei~Zhang, Ziyao~Tang, Yuejia~Ma, Yuanzhe~Chen, Shuchao~Song, and Boshi~Tang%
\thanks{Jiayuan Gao, Yingwei Zhang, and Shuchao Song are with the Beijing Key Laboratory of Mobile Computing and Pervasive Device, Institute of Computing Technology, Chinese Academy of Sciences, Beijing, China, and also with the University of Chinese Academy of Sciences, Beijing, China (e-mail: gaojiayuan20z@ict.ac.cn; zhangyingwei@ict.ac.cn; songshuchao22b@ict.ac.cn).}%
\thanks{Ziyao Tang is with Nanyang Technological University, Singapore (e-mail: ziyao008@e.ntu.edu.sg).}%
\thanks{Yuejia Ma is with the University of Science and Technology Beijing, Beijing, China (e-mail: u202341643@xs.ustb.edu.cn).}%
\thanks{Yuanzhe Chen is with the University of Chinese Academy of Sciences, Beijing, China (e-mail: chenyuanzhe21s@ict.ac.cn).}%
\thanks{Boshi Tang is with Tsinghua University, Beijing, China (e-mail: tbs16@mails.tsinghua.edu.cn).}%
\thanks{Corresponding author: Yingwei Zhang.}}

\markboth{IEEE Transactions on Mobile Computing,~Vol.~XX, No.~X, 2026}%
{Gao \MakeLowercase{\textit{et al.}}: Coverage-Aware Virtual IMU Augmentation}

\maketitle

\begin{center}
\footnotesize
This work has been submitted to the IEEE for possible publication.
Copyright may be transferred without notice, after which this version
may no longer be accessible.
\end{center}

\begin{abstract}
IMU-based human activity recognition (HAR) enables continuous, privacy-friendly monitoring of daily activities using wearable sensors.
However, building reliable HAR models that generalize across diverse users and real-world conditions requires large amounts of labeled IMU data, which are expensive and difficult to collect.
Existing approaches mainly rely on augmentation or synthesis to expand available data, but indiscriminately adding virtual samples may provide little new coverage and introduce unreliable supervision.
To overcome these challenges, we propose a novel coverage-aware virtual IMU augmentation framework that decides where to supplement real data, how to generate and select virtual candidates, and how strongly to weight them during training.
Specifically, we select diversity and scarcity anchors in a learned sensor embedding space, convert anchor dynamics into prompts, and generate virtual IMU candidates for each anchor. We then rank candidates by a selection cost combining anchor proximity and label consistency, and incorporate the selected candidates into HAR training with reliability-based weights.
Experiments on public HAR benchmarks show that our method consistently improves recognition performance over competitive baselines, and ablation studies confirm the effectiveness of the proposed framework design.
\end{abstract}


\section{Introduction}


IMU-based human activity recognition (HAR) recognizes daily activities from signals collected by wearable inertial measurement units~\cite{xu2023practically}.
It supports applications such as daily activity monitoring, fitness assistance, and home-based rehabilitation~\cite{wang2024ubiphysio}.
Compared with camera-based systems, wearable IMUs reduce visual privacy concerns and support continuous monitoring in everyday environments~\cite{xie2025harmony}.
However, reliable HAR models require training data that capture important variations across users, activity execution styles, sensor placements, and sensing environments.
A finite labeled dataset only provides an incomplete view of the activity distribution, and collecting data across these sources of variation is costly and often constrained by time, supervision, and deployment conditions.

\begin{figure}[t]
\centering
\includegraphics[width=\linewidth]{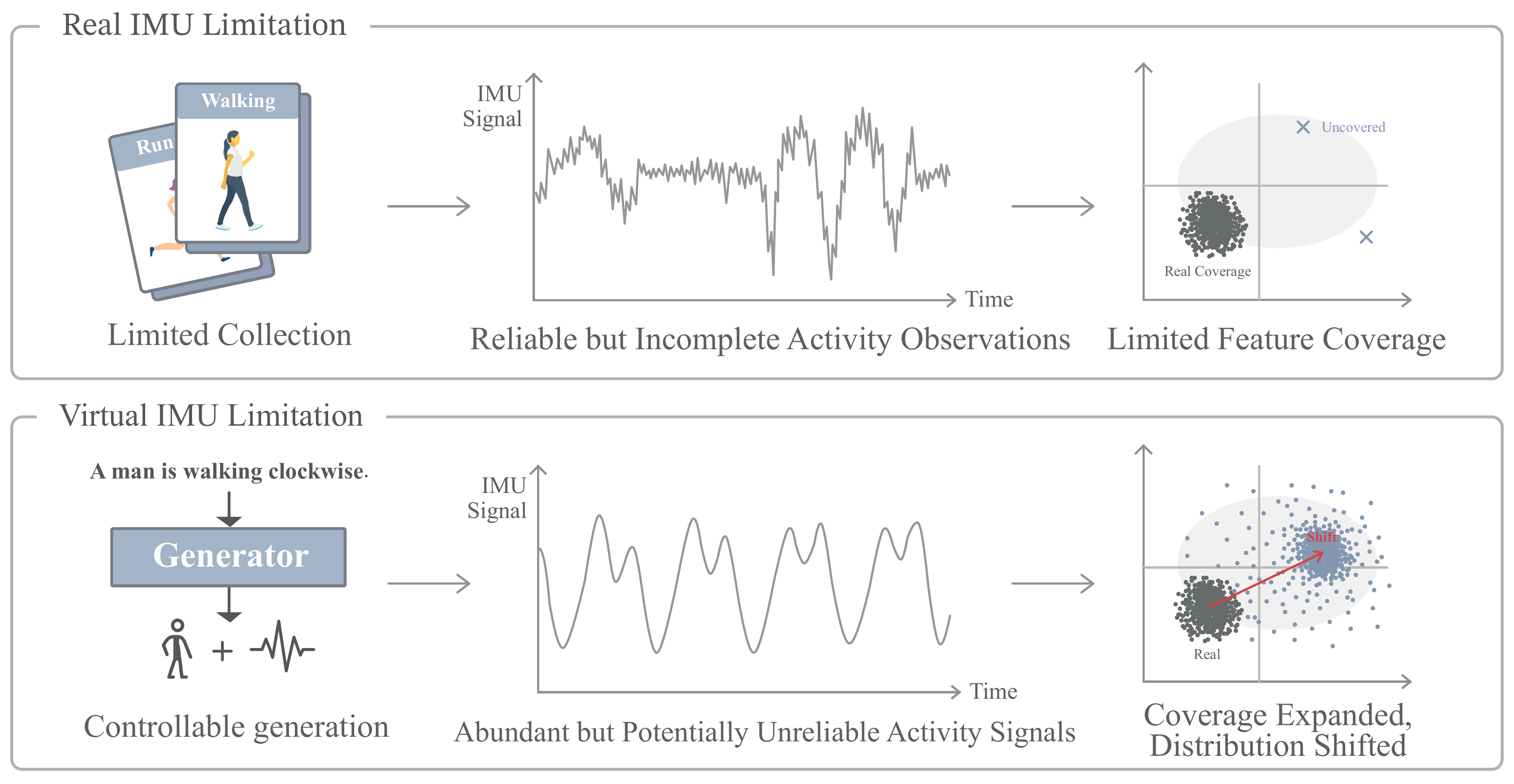}
\caption{Motivation for coverage-aware virtual IMU augmentation. Limited real IMU collections provide reliable but incomplete observations of activity patterns, whereas virtual IMU data can expand training coverage but may introduce shifted or unreliable samples.}
\label{fig:pic1}
\end{figure}


As illustrated in Fig.~\ref{fig:pic1}, virtual IMU data can expand training coverage beyond what a limited real IMU collection can provide.
Since such data can be generated at scale and conditioned on target activities, they provide a practical way to supplement limited real training data~\cite{leng2023generating,leng2024imugpt}.
Recent studies on diffusion-based dataset distillation also suggest that synthetic samples should be evaluated not only by realism but also by their downstream utility, representativeness, diversity, and artifact mitigation~\cite{chen2025influence,chan2025mgd}.
This issue is particularly important for IMU-based HAR. Even a seemingly reasonable generated sequence may provide little additional coverage if it falls in a region already well represented by real training data, or it may deviate from activity-specific patterns. Treating such samples the same as real data can degrade model performance.
Therefore, we treat virtual IMU samples as candidate training data: their utility depends on the coverage they provide and their consistency with the target activity, while their influence during training is determined by their reliability.

To address these challenges, we propose a coverage-aware virtual IMU augmentation framework for HAR with limited real training data.
The framework addresses three practical questions: \emph{where to supplement the real training distribution}, \emph{how to generate virtual candidates}, and \emph{how to control their influence during training}.

First, we select real training windows as anchors in a learned sensor embedding space.
Diversity anchors represent distinct within-class motion patterns, while scarcity anchors are selected using a local-density criterion to target sparsely covered regions within the observed training distribution.
Second, each anchor is converted into an anchor-conditioned prompt using anchor-level dynamics attributes such as tempo, intensity, and periodicity, and multiple virtual IMU candidates are generated for each anchor.
Third, the generated candidates are assessed using their embedding-space proximity to the corresponding anchor and label consistency estimated by the classifier head of a seed encoder trained only on real training data.
Based on this assessment, selected candidates are incorporated into HAR training with reliability-based sample weights, so that less reliable virtual samples contribute less to model optimization.


The main contributions are summarized as follows:
\begin{enumerate}
\item We propose a coverage-aware virtual IMU augmentation framework for HAR with limited real training data, where anchors selected from the real training data guide virtual IMU generation, and the resulting candidates are assessed to determine which are retained and how strongly they contribute to HAR training.

\item We introduce an anchor-conditioned virtual IMU generation strategy. It selects real training windows as diversity and scarcity anchors in a learned sensor embedding space and constructs generation prompts from anchor-level dynamics attributes, thereby guiding virtual candidate generation toward diverse within-class patterns and sparsely covered regions of the observed training distribution.

\item We propose a reliability-based candidate selection and weighting scheme.
It computes selection costs from embedding-space proximity to the corresponding anchor and label consistency, retains the candidates with the lowest selection costs, and incorporates them into HAR training with reliability-based sample weights.
\end{enumerate}

The remainder of this paper is organized as follows.
Section~2 reviews related work on virtual and synthetic data for activity recognition, label-efficient learning and domain generalization, and noisy-label learning and robust training.
Section~3 introduces the problem setup and presents the proposed framework for anchor-conditioned virtual IMU generation, candidate selection, and reliability-weighted integration.
Section~4 describes the experimental setup and reports the evaluation results. Section~5 discusses limitations and future work. Section~6 concludes the paper.

\section{Related Work}
\subsection{Virtual and Synthetic Data for Activity Recognition}

Virtual and synthetic data are widely used to mitigate labeled data scarcity in human activity recognition.
Existing approaches can be broadly categorized into four directions: signal-level augmentation, generative time-series modeling, text-driven virtual IMU generation, and model-based IMU simulation.
Signal-level methods apply transformations to temporal structure, amplitude, or orientation, while recent work explores automated augmentation policy search and physically plausible augmentation strategies~\cite{zhou2024autoaughar,leng2025scaling,oishi2026physically}.
Generative models synthesize inertial sequences for low-resource settings, with recent work exploring diffusion-based time-series generation~\cite{leng2025scaling,oppel2025diffusion}.
Text-driven methods generate virtual IMU signals from textual descriptions through cross-modal motion synthesis pipelines~\cite{leng2023generating,leng2024imugpt,haeusler2025text2imu}.
Meanwhile, model-based simulation approaches use human-body and sensor models to generate synthetic inertial data with explicit control over motion and sensor configurations~\cite{uhlenberg2024synhar,oishi2025wimusim,oishi2026physically}.

Overall, these approaches have advanced synthetic inertial data generation in terms of scale, diversity, realism, and physical plausibility across augmentation, generative, and simulation paradigms~\cite{leng2025scaling,leng2024imugpt,uhlenberg2024synhar,oishi2025wimusim,oppel2025diffusion,oishi2026physically}.
Recent studies on diffusion-based dataset distillation further suggest that synthetic data should be evaluated not only by realism but also by downstream utility, diversity, and artifact mitigation~\cite{chen2025influence,chan2025mgd}.
However, the above augmentation and generation methods are primarily designed to optimize transformation policies, generative fidelity, or physical consistency, without explicitly considering which regions of the real training distribution are well represented or underrepresented.
Consequently, synthetic samples may satisfy their method-specific objectives while providing little additional coverage of underrepresented regions in the empirical class-conditional distribution.
This motivates a coverage-aware view of virtual IMU generation that considers both the empirical training distribution and the reliability of generated samples during downstream learning.

\subsection{Label-Efficient Learning and Domain Generalization}

Label-efficient learning in sensor-based human activity recognition aims to reduce annotation costs by exploiting unlabeled or sparsely labeled sensor streams.
Semi-supervised approaches improve HAR models by jointly using labeled and unlabeled samples, including through interpolation-based training~\cite{duan2024wearable}.
Self-training methods further exploit unlabeled sequences via pseudo-labeling, iteratively refining models with high-confidence predictions~\cite{tang2021selfhar}.
In addition, unsupervised cross-user domain adaptation transfers knowledge from labeled source subjects to unlabeled target subjects, addressing user variation without target-domain labels~\cite{hu2023swl}.

Beyond annotation efficiency, recent domain-general HAR studies focus on robustness under distribution shift.
Invariant representation learning has been studied under cross-subject, cross-dataset, and cross-position settings~\cite{xiong2025generalizable}, while recent benchmarks evaluate the generalization of self-supervised HAR models to unseen target distributions~\cite{cai2026benchhar}.
Together, these studies highlight the importance of learning sensor representations that generalize across users, datasets, and sensor positions.

However, the above label-efficient and domain-general HAR methods primarily improve learning from the available observations.
They improve data utilization and representation transfer, but do not explicitly expand the empirical coverage of underrepresented activity patterns.
Therefore, they do not address how generated sensor data can complement limited real-data coverage without introducing unreliable training signals.

\subsection{Noisy-Label Learning and Robust Training}

Noisy-label learning and robust training aim to reduce the effect of corrupted labels by making optimization less sensitive to unreliable supervision.
Representative approaches include robust loss functions such as the generalized cross-entropy loss, which reduces sensitivity to incorrect labels~\cite{zhang2018generalized}.
Another line of work focuses on sample selection and reweighting, where small-loss or high-confidence instances are prioritized during training.
Co-teaching improves robustness by training two networks that select small-loss samples for each other, thereby reducing error accumulation from noisy labels~\cite{han2018co}.
Subsequent methods extend this line through instance-specific sample selection, label correction, and confidence-aware regularization or tracking~\cite{li2023disc,zhang2023rankmatch,sheng2024foster,pan2025enhanced,yuan2025enhancing}.

These studies show that treating all training samples as equally reliable can degrade learning, and that confidence-aware or selection-based strategies can mitigate noise-induced error propagation.
However, they are mainly designed for fixed training sets with potentially corrupted labels, rather than for settings where additional training samples are produced by a generative process.
For virtual IMU data, unreliability may arise from generation artifacts, distribution mismatch, or inconsistency with the intended activity.
Accordingly, generated sequences should not be treated as uniformly reliable training data, and their contribution to learning should reflect their estimated reliability.

\section{Methodology}

\begin{figure*}[t]
\centering
\includegraphics[width=\textwidth]{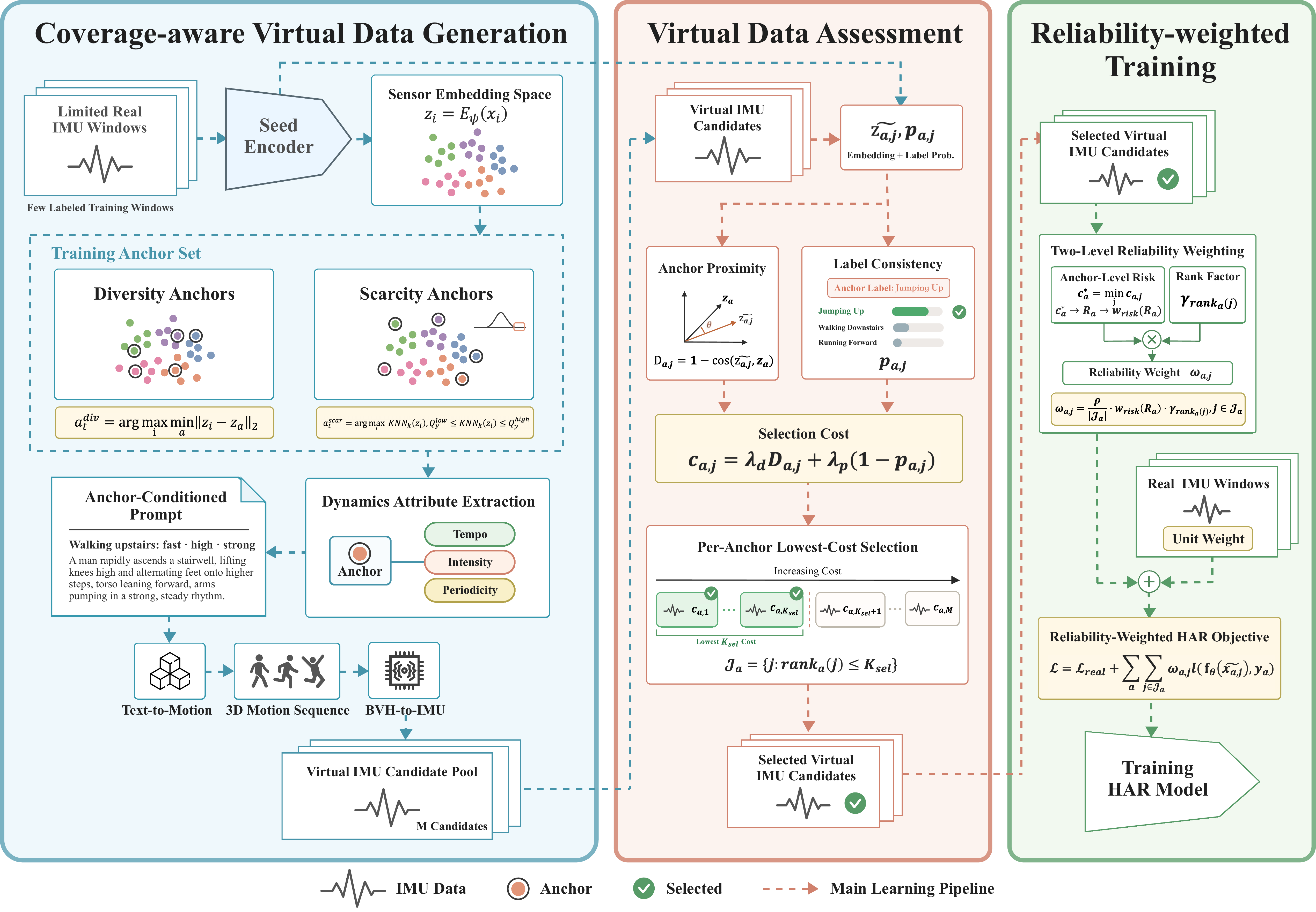}
\caption{\textbf{Framework overview.}
(1) \textbf{Anchor-guided virtual IMU generation:}
select real training windows as diversity and scarcity anchors in a learned sensor embedding space, construct anchor-conditioned prompts from activity labels and anchor-level tempo, intensity, and periodicity attributes, and synthesize multiple virtual IMU candidates.
(2) \textbf{Virtual candidate assessment:}
assess candidates using their embedding-space proximity to the corresponding anchors and label consistency estimated by the seed classifier head, combine the two criteria into a selection cost, and retain the candidates with the lowest costs for each anchor.
(3) \textbf{Reliability-weighted HAR training:}
assign reliability-based sample weights using candidate selection costs and within-anchor ranks, and train the HAR model on real IMU windows and selected virtual candidates.}
\label{fig:pic2}
\end{figure*}

The proposed framework is summarized in Fig.~\ref{fig:pic2}.
Building robust sensor-based HAR models is challenging because labeled IMU data are costly to obtain and activity signals vary across users, execution styles, and sensing conditions~\cite{chen2021deep}.
A finite real collection therefore provides an incomplete empirical view of the activity distribution.
Virtual IMU generation can broaden this empirical coverage, but generated samples should not be assumed to be as reliable as real labeled samples.
We address this issue by grounding virtual IMU generation in real training anchors and integrating generated samples according to their estimated reliability.
Specifically, we select real training windows as anchors using a learned sensor embedding space, convert their dynamics attributes into prompts, and generate a pool of virtual IMU candidates.
The candidates are assessed by their embedding-space proximity to the corresponding anchors and label consistency, and the selected candidates are integrated into HAR training with reliability-based weights.

We first formalize the window-level HAR setting and the leave-one-subject-out (LOSO) evaluation protocol in Sec.~\ref{sec:setup}.
We then introduce anchor-conditioned virtual IMU generation in Sec.~\ref{sec:semantic_data_generation}, virtual candidate assessment and selection in Sec.~\ref{sec:quality_assess}, and reliability-weighted integration in Sec.~\ref{sec:risk_erm}.
Algorithm~\ref{alg:anchor_guided_training} summarizes the complete procedure.

\subsection{Problem Setup}
\label{sec:setup}
We formulate IMU-based HAR as window-level multi-class classification: a model receives a multichannel IMU time window and predicts its activity label. Let $\mathcal{U}=\{1,\ldots,S\}$ denote the set of subjects.
For each subject $m\in\mathcal{U}$, let $\mathcal{D}_m$ denote the set of labeled real IMU windows:
\begin{equation}
\mathcal{D}_m
=
\{(x_{m,i},y_{m,i})\}_{i=1}^{N_m},
\qquad
x_{m,i}\in\mathbb{R}^{T\times d},
\quad
y_{m,i}\in\mathcal{Y}.
\label{eq:subject_windows}
\end{equation}
Here $x_{m,i}$ is the $i$-th IMU window from subject $m$, $y_{m,i}$ is its activity label, $T$ is the window length, $d$ is the number of IMU channels, and $\mathcal{Y}$ is the activity label set. The subject index is used only to define cross-subject splits and is not part of the model input.

We evaluate under the LOSO protocol. For a held-out subject $u$, the training and test sets are
\begin{equation}
\mathcal{D}_{\mathrm{tr}}^{(u)}
=
\bigcup_{m\in\mathcal{U}\setminus\{u\}}\mathcal{D}_m,
\qquad
\mathcal{D}_{\mathrm{te}}^{(u)}
=
\mathcal{D}_u.
\label{eq:loso_split}
\end{equation}
All fold-specific components are constructed without access to $\mathcal{D}_{\mathrm{te}}^{(u)}$.
These include seed encoder training, anchor selection, dynamics-attribute statistics, candidate assessment, weight assignment, and normalization statistics.
The held-out set $\mathcal{D}_{\mathrm{te}}^{(u)}$ is used only for final evaluation on real IMU windows.

For notational simplicity, we re-index the real training windows in $\mathcal{D}_{\mathrm{tr}}^{(u)}$ as ${(x_i,y_i)}$ when the subject index is not needed.

\subsection{Anchor-Conditioned Virtual IMU Generation}
\label{sec:semantic_data_generation}
A finite IMU training set provides only a partial empirical view of the within-class variation of each activity.
To address this issue, we select real training windows as generation anchors based on their positions within each class in a learned embedding space.
Specifically, we train a seed network on the real IMU windows in the current LOSO training split and use the resulting embedding space to select two complementary types of anchors.
Diversity anchors represent distinct within-class motion patterns, whereas scarcity anchors target sparsely covered regions while excluding extreme outliers.
Each anchor provides a real sensor reference and the dynamics attributes used to construct its generation prompt.

For each real training window $(x_i,y_i)\in\mathcal{D}_{\mathrm{tr}}^{(u)}$, we compute the seed embedding
\begin{equation}
z_i=E_{\psi}(x_i).
\label{eq:seed_representation}
\end{equation}
Here $E_{\psi}$ and $H_{\psi}$ denote the encoder and classifier head, respectively, of the seed network trained on $\mathcal{D}_{\mathrm{tr}}^{(u)}$, and $z_i$ is the embedding of window $x_i$.
The classifier head $H_{\psi}$ is later used to assess the label consistency of generated candidates.

For each class $y\in\mathcal{Y}$, let $\mathcal{I}_y^{(u)}$ denote the indices of its real training windows in $\mathcal{D}_{\mathrm{tr}}^{(u)}$. The full anchor set for split $u$ is
\begin{equation}
\mathcal{A}^{(u)}
=
\bigcup\limits_{y\in\mathcal{Y}}
\bigl(
\mathcal{A}_y^{\mathrm{div}}
\cup
\mathcal{A}_y^{\mathrm{scar}}
\bigr).
\label{eq:anchor_set}
\end{equation}
We represent each anchor by the index of its corresponding real training window; accordingly, $x_a$, $y_a$, and $z_a$ denote the anchor window, label, and embedding, respectively.
For each class, diversity-anchor selection is initialized with the smallest-index training sample of that class, with $a_{y,1}^{\mathrm{div}}=\min \mathcal{I}_y^{(u)}$ and
$\mathcal{A}_{y,1}^{\mathrm{div}}=\{a_{y,1}^{\mathrm{div}}\}$.
The remaining anchors are then selected using class-wise greedy farthest-point sampling:
\begin{equation}
a_{y,t}^{\mathrm{div}}
=
\arg\max_{i\in
\mathcal I_y^{(u)}\setminus
\mathcal A_{y,t-1}^{\mathrm{div}}}
\min_{a\in\mathcal A_{y,t-1}^{\mathrm{div}}}
\lVert z_i-z_a\rVert_2
\qquad
t=2,\ldots,K_{\mathrm{div}}.
\label{eq:diversity_anchor}
\end{equation}
Here, $\mathcal A_{y,t-1}^{\mathrm{div}}$ contains the diversity anchors selected before iteration $t$. The procedure yields $K_{\mathrm{div}}$ diversity anchors for each class.

Scarcity anchors are selected by a class-conditional local-density criterion:
\begin{equation}
\delta_i^{\mathrm{scar}}
=
d_k\!\left(z_i;\mathcal{I}_y^{(u)}\right),
\qquad
Q_y^{\mathrm{low}}
\le
\delta_i^{\mathrm{scar}}
\le
Q_y^{\mathrm{high}}.
\label{eq:scarcity_anchor}
\end{equation}
Here, $d_k\!\left(z_i;\mathcal{I}_y^{(u)}\right)$ is the Euclidean distance from $z_i$ to its $k$-th nearest same-class neighbor, excluding itself, and $Q_y^{\mathrm{low}}$ and $Q_y^{\mathrm{high}}$ are class-wise empirical quantile thresholds computed within the current training fold.
Samples satisfying this interval are treated as scarcity candidates.
For each class, we retain up to $K_{\mathrm{scar}}$ candidates with the largest sparsity scores to form $\mathcal{A}_y^{\mathrm{scar}}$.
The lower threshold focuses selection on sparsely covered regions, while the upper threshold excludes extreme outliers.

Real IMU anchors provide low-level sensor dynamics, whereas text-conditioned motion generators operate on natural-language motion descriptions.
To bridge this gap, we express the anchor dynamics as textual motion cues and combine them with the activity label to construct an anchor-conditioned prompt.
For each anchor $a$, we extract a compact dynamics descriptor from the anchor window:
\begin{equation}
\begin{aligned}
\mathbf{b}_a
&=
\Phi(x_a)
=
\bigl(
b_a^{\mathrm{tempo}},
b_a^{\mathrm{intensity}},
b_a^{\mathrm{periodicity}}
\bigr),\\
\tau_a
&=
\operatorname{Prompt}(y_a,\mathbf{b}_a).
\end{aligned}
\label{eq:anchor_prompt}
\end{equation}
The anchor dynamics are described by three attributes:
\begin{itemize}[leftmargin=*]
\item \textbf{Tempo} describes the movement rate of the anchor, indicating whether the target motion is slow, moderate, or fast.
\item \textbf{Intensity} reflects the inertial energy in the accelerometer and gyroscope channels and represents the magnitude of movement.
\item \textbf{Periodicity} characterizes repeated temporal structure, distinguishing stable rhythmic patterns from weakly periodic or irregular motion.
\end{itemize}
Here, $\Phi(\cdot)$ extracts the dynamics descriptor $\mathbf{b}_a$ from the anchor window, and $\operatorname{Prompt}$ maps $(y_a,\mathbf{b}_a)$ to an anchor-conditioned motion prompt $\tau_a$.
In implementation, each prompt contains one activity field and three anchor-derived dynamics fields: tempo, intensity, and periodicity.
For example, an anchor labeled ``walking forward'' with medium tempo, low intensity, and strong periodicity is represented by a prompt describing walking at a moderate tempo with low movement intensity and a regular rhythm.
Finally, a fixed synthesis interface converts the prompt into virtual IMU candidates:
\begin{equation}
\{\tilde{x}_{a,j}\}_{j=1}^{M}
=
G(\tau_a),
\qquad
a\in\mathcal{A}^{(u)}.
\label{eq:virtual_candidates}
\end{equation}
Here, $G$ denotes the fixed synthesis interface, $\tilde{x}_{a,j}$ is the $j$-th virtual IMU candidate generated for anchor $a$, and $M$ is the candidate budget per anchor.
We instantiate $G$ with the synthesis pipeline adopted by IMUGPT~\cite{leng2023generating,leng2024imugpt}, using a pretrained T2M-GPT model~\cite{zhang2023generating} followed by IMUSim~\cite{young2011imusim}.
T2M-GPT maps $\tau_a$ to a 3D body-motion sequence, which is converted to BVH through inverse kinematics.
IMUSim then simulates ideal accelerometer and gyroscope signals at the skeleton joint nearest to the target sensor location.
The synthesis backend is kept fixed throughout all experiments.

\subsection{Virtual Candidate Assessment and Selection}
\label{sec:quality_assess}
Virtual IMU candidates are generated to supplement motion patterns that are sparsely represented in the real training data.
However, apparent realism alone does not ensure that a candidate is suitable for HAR training.
In IMU-based HAR, generated signals that fail to preserve activity-consistent inertial patterns may introduce label noise and degrade downstream training.
We therefore assess and select generated candidates before adding them to the training set.
The assessment uses the corresponding anchor embeddings and the seed network trained within the current LOSO training split.
We select candidates that remain close to their corresponding real anchors in the embedding space and receive high predicted probabilities for the intended activity labels.

For each generated candidate $\tilde{x}_{a,j}$, we compute its embedding and the predicted probability of the intended activity label using the seed network:

\begin{equation}
\tilde{z}_{a,j}=E_{\psi}(\tilde{x}_{a,j}),
\qquad
p_{a,j}=\left[H_{\psi}(\tilde{z}_{a,j})\right]_{y_a}.
\label{eq:candidate_representation}
\end{equation}
Here, $\tilde{z}_{a,j}$ is the candidate embedding, and $p_{a,j}$ is the probability assigned to the intended activity label $y_a$.

We quantify anchor proximity using the cosine distance between the candidate and anchor embeddings:
\begin{equation}
D_{a,j}^{\mathrm{emb}}
=
1-
\frac{
\mathbf{z}_a^{\top}\tilde{\mathbf{z}}_{a,j}
}{
\lVert \mathbf{z}_a\rVert_2
\lVert \tilde{\mathbf{z}}_{a,j}\rVert_2
}.
\label{eq:candidate_assessment_terms}
\end{equation}
A smaller $D_{a,j}^{\mathrm{emb}}$ indicates greater similarity to the corresponding anchor, while a larger $p_{a,j}$ indicates stronger support for the intended activity label.

The candidate selection cost is defined as
\begin{equation}
c_{a,j}=\lambda_dD_{a,j}^{\mathrm{emb}}
+
\lambda_p(1-p_{a,j}),
\qquad
\lambda_d,\lambda_p\ge 0,
\quad
\lambda_d+\lambda_p=1.
\label{eq:candidate_selection_cost}
\end{equation}
Lower values of $c_{a,j}$ indicate candidates that remain closer to the anchor representation and receive stronger support for the intended activity label.

For each anchor, we retain the candidates with the lowest selection costs. Let $\operatorname{rank}_a(j)$ denote the rank of candidate $j$ when the costs ${c_{a,l}}$ are sorted in ascending order. The selected candidate indices are:
\begin{equation}
\mathcal{J}_a=\{
j\in\{1,\ldots,M\}
:
\operatorname{rank}_a(j)
\le
K_{\mathrm{sel}}
\}
\label{eq:selected_virtual_set}
\end{equation}
Here, $K_{\mathrm{sel}}$ is the maximum number of candidates retained for each anchor, and $\mathcal{J}_a$ is the corresponding set of selected candidate indices.
The selected candidates are subsequently integrated into HAR training using reliability-based weights.

\subsection{Reliability-Weighted Integration}
\label{sec:risk_erm}
The selected virtual candidates are not treated as equally reliable.
We assign weights at two levels: an anchor-level risk factor determined by the best candidate cost for each anchor, and a candidate-level rank factor determined by the within-anchor cost ordering.
For each anchor, we summarize candidate reliability using its lowest selection cost:
\begin{equation}
c_a^\star
=
\min_{j=1,\ldots,M} c_{a,j}.
\label{eq:anchor_best_cost}
\end{equation}
A lower $c_a^\star$ indicates that the synthesis process produces at least one candidate with strong anchor proximity and label consistency.

We divide anchors into low, medium, and high risk groups using the lower and upper empirical tertiles of
$\{c_a^\star:a\in\mathcal A^{(u)}\}$
within the current training fold.
Let $\tau_{\mathrm{low}}^{(u)}$ and $\tau_{\mathrm{high}}^{(u)}$ denote these fold-specific thresholds.
The risk category of anchor $a$ is
\begin{equation}
R_a
=
\begin{cases}
\mathrm{L},
&
c_a^\star\le\tau_{\mathrm{low}}^{(u)},\\
\mathrm{M},
&
\tau_{\mathrm{low}}^{(u)}
<
c_a^\star
\le
\tau_{\mathrm{high}}^{(u)},\\
\mathrm{H},
&
c_a^\star>\tau_{\mathrm{high}}^{(u)}.
\end{cases}
\label{eq:anchor_risk}
\end{equation}

We map the anchor-level risk categories to discrete reliability factors:
\begin{equation}
w_{\mathrm{risk}}(R_a)
=
\begin{cases}
w_{\mathrm L}, & R_a=\mathrm L,\\
w_{\mathrm M}, & R_a=\mathrm M,\\
w_{\mathrm H}, & R_a=\mathrm H,
\end{cases}
\qquad
w_{\mathrm L}\ge w_{\mathrm M}\ge w_{\mathrm H}\ge 0.
\label{eq:risk_mapping}
\end{equation}
For each selected candidate $j\in\mathcal{J}_a$, the final training weight is

\begin{equation}
\omega_{a,j}
=
\frac{\rho}{|\mathcal{J}_a|}
w_{\mathrm{risk}}(R_a)
\gamma_{\operatorname{rank}_a(j)},
\qquad
j\in\mathcal{J}_a.
\label{eq:training_weight}
\end{equation}
Here, $\rho$ controls the overall contribution of virtual samples, and
$\gamma_{\operatorname{rank}_a(j)}$ determines the relative contribution of candidates within the same anchor, with
\begin{equation}
1=\gamma_1
\ge
\gamma_2
\ge
\cdots
\ge
\gamma_{K_{\mathrm{sel}}}
\ge 0.
\label{eq:rank_weights}
\end{equation}
The factor $1/|\mathcal{J}_a|$ prevents the aggregate contribution of an anchor from growing linearly with the number of selected candidates.
The risk factor is shared by candidates generated from the same anchor, while the rank factor assigns greater influence to lower-cost candidates within that anchor.

After weight assignment, the weighted virtual set for split $u$ is defined as
\begin{equation}
\mathcal{V}^{(u)}
=
\{(\tilde{x}_{a,j},y_a,\omega_{a,j}):
a\in\mathcal{A}^{(u)},\,j\in\mathcal{J}_a\}.
\label{eq:weighted_virtual_set}
\end{equation}

To assign unit weight to real samples, we train the HAR model $f_\theta$ using the weighted empirical risk minimization objective.
\begin{equation}
\begin{aligned}
\mathcal{L}(\theta)
&=
\sum_{(x_i,y_i)\in\mathcal{D}_{\mathrm{tr}}^{(u)}}
\ell\bigl(f_\theta(x_i),y_i\bigr)\\
&\quad+
\sum_{a\in\mathcal{A}^{(u)}}
\sum_{j\in\mathcal{J}_a}
\omega_{a,j}
\ell\bigl(f_\theta(\tilde{x}_{a,j}),y_a\bigr).
\end{aligned}
\label{eq:reliability_weighted_objective}
\end{equation}
Here, $\ell$ denotes the supervised classification loss.
All virtual samples, weights, and selection costs are computed strictly from the training split, without access to the held-out subject during generation, selection, weighting, or optimization.

\begin{algorithm2e}[!h]
\DontPrintSemicolon
\footnotesize
\caption{Coverage-Aware Virtual IMU Augmentation for HAR.}
\label{alg:anchor_guided_training}

\KwIn{Training set $\mathcal{D}_{\mathrm{tr}}^{(u)}$,
fixed synthesis interface $G$, and the anchor-selection,
candidate-selection, and integration hyperparameters}

\KwOut{Trained HAR model $f_\theta$}

Train the seed network $(E_\psi,H_\psi)$ on
$\mathcal{D}_{\mathrm{tr}}^{(u)}$ and compute
$z_i=E_\psi(x_i)$ for all real training windows\;

For each class, select $K_{\mathrm{div}}$ diversity anchors by
greedy farthest-point sampling and up to $K_{\mathrm{scar}}$
scarcity anchors by the class-conditional sparsity criterion\;

\ForEach{anchor $a\in\mathcal{A}^{(u)}$}{
  $\mathbf{b}_a\leftarrow\Phi(x_a)$\;
  $\tau_a\leftarrow
  \operatorname{Prompt}(y_a,\mathbf{b}_a)$\;
  $\{\tilde{x}_{a,j}\}_{j=1}^{M}
  \leftarrow G(\tau_a)$\;

  \For{$j=1,\ldots,M$}{
    Compute $\tilde{z}_{a,j}$, $p_{a,j}$,
    $D_{a,j}^{\mathrm{emb}}$, and $c_{a,j}$
    using Eqs.~\eqref{eq:candidate_representation}--%
    \eqref{eq:candidate_selection_cost}\;
  }

  Rank candidates in ascending order of $c_{a,j}$ and let
  $\operatorname{rank}_a(j)$ denote the within-anchor rank\;

  $\mathcal{J}_a
  \leftarrow
  \{j:\operatorname{rank}_a(j)\le K_{\mathrm{sel}}\}$\;

  $c_a^\star
  \leftarrow
  \min_{j=1,\ldots,M}c_{a,j}$\;
}

Compute $\tau_{\mathrm{low}}^{(u)}$ and
$\tau_{\mathrm{high}}^{(u)}$ as the lower and upper empirical
tertiles of $\{c_a^\star:a\in\mathcal{A}^{(u)}\}$\;

$\mathcal{V}^{(u)}\leftarrow\emptyset$\;

\ForEach{anchor $a\in\mathcal{A}^{(u)}$}{
  Assign $R_a$ according to Eq.~\eqref{eq:anchor_risk}\;

  \ForEach{$j\in\mathcal{J}_a$}{
    Compute $\omega_{a,j}$ using
    Eq.~\eqref{eq:training_weight}\;

    Add $(\tilde{x}_{a,j},y_a,\omega_{a,j})$
    to $\mathcal{V}^{(u)}$\;
  }
}

Train $f_\theta$ on $\mathcal{D}_{\mathrm{tr}}^{(u)}$
and $\mathcal{V}^{(u)}$ using
Eq.~\eqref{eq:reliability_weighted_objective}\;

\Return{$f_\theta$}\;
\end{algorithm2e}

\section{Experiments}
\subsection{Experimental Setup}
\subsubsection{Dataset}

We evaluate the proposed method on three IMU-based HAR benchmarks: USC-HAD~\cite{zhang2012usc}, PAMAP2~\cite{reiss2012introducing}, and RealWorld~\cite{sztyler2016body}.
We further evaluate the method in an exploratory medical IMU classification setting using PADS~\cite{varghese2024machine,PhysioNet-parkinsons-disease-smartwatch-1.0.0,brenner2022reducing}.
USC-HAD contains daily activities recorded using a single IMU worn at the front-right hip.
PAMAP2 provides recordings of diverse physical activities from multiple wearable sensors; in our experiments, we use the chest IMU recordings.
RealWorld captures wearable activity data in less constrained daily environments, exhibiting stronger real-world sensing variability; we use the forearm IMU recordings in our experiments.
PADS contains wrist-worn IMU recordings from clinical movement assessments and is treated as an exploratory medical extension rather than a standard HAR benchmark.
Table~\ref{tab:dataset-stats} summarizes the dataset scope and evaluation protocols.

\begin{table}[t]
\centering
\caption{Experimental datasets, sensor configurations, and evaluation protocols.}
\label{tab:dataset-stats}
\scriptsize
\setlength{\tabcolsep}{2pt}
\renewcommand{\arraystretch}{1.12}
\begin{tabular}{lcccc}
\hline
\multicolumn{5}{c}{\textbf{HAR benchmarks}} \\
\hline
Dataset & Subjects & Activities & Sensor placement & Protocol \\
\hline
USC-HAD   & 14 & 7 & front-right hip & LOSO \\
PAMAP2    & 8  & 5 & chest & LOSO \\
RealWorld & 15 & 5 & forearm & LOSO \\
\hline
\multicolumn{5}{c}{\textbf{Exploratory medical extension}} \\
\hline
Dataset & Folds & Task & Sensor placement & Protocol \\
\hline
PADS & 5 & PD-vs-HC classification & wrist & subject-wise split \\
\hline
\end{tabular}
\end{table}

For USC-HAD, PAMAP2, and RealWorld, we evaluate dynamic activity subsets under the LOSO protocol to assess cross-subject generalization.
We focus on dynamic activities because their temporal dynamics and signal-amplitude variations make them more suitable than near-static postures for evaluating virtual IMU generation.
In each LOSO fold, all windows from one subject form the test set, while the remaining subjects provide the data used for training and any fold-specific model selection.
Virtual IMU data are used only for training.
The held-out subject is never used for normalization, virtual candidate construction, or model fitting within each LOSO fold.
LOSO thus evaluates generalization under subject-level distribution shift.
For these HAR benchmarks, we segment IMU streams using a sliding window of 2s with a step size of 1s.
All real IMU streams are downsampled to 20 Hz to match the virtual data sampling rate, and each channel is normalized using statistics computed from the training subjects in the current fold.
The same preprocessing pipeline is used across all compared methods.
For PADS, we use the five-fold subject-wise split reported in Table~\ref{tab:dataset-stats} and report the results separately as an exploratory medical extension.

\subsubsection{Evaluation Metrics}

For USC-HAD, PAMAP2, and RealWorld, we evaluate performance on the held-out subject in each LOSO fold using Macro-F1 as the primary metric and overall accuracy (Acc) as a secondary metric.
Macro-F1 is used as the primary metric because it weights all activity classes equally and reduces the dominance of majority classes, whereas accuracy summarizes overall recognition performance.
We first compute metrics on each LOSO fold, average them across folds for each seed, and then report the mean $\pm$ standard deviation across seeds.
All methods use identical dataset splits and evaluation procedures within each benchmark to ensure fair comparison.
For PADS, we report Macro-F1 and balanced accuracy separately as an exploratory medical extension.

\subsubsection{Implementation Details}

We evaluate three downstream HAR models: DeepConvLSTM~\cite{ordonez2016deep}, DeepConvLSTM-Attention~\cite{murahari2018attention}, and MLP-HAR~\cite{zhou2024mlp}.
DeepConvLSTM combines convolutional feature extraction with recurrent temporal modeling, while DeepConvLSTM-Attention further incorporates temporal attention pooling.
MLP-HAR serves as a lightweight feed-forward alternative.
Within each benchmark and downstream model, all compared training configurations use the same splits, preprocessing, and model configuration to ensure fair comparison.

Reliability-based weighting is applied only during training.
Real samples are assigned unit weight, while selected virtual IMU samples are weighted according to the anchor-level risk category and their within-anchor ranks.
For classical classifiers, weights are incorporated through sample-weighted optimization; for neural models, they are applied as per-sample loss weights.
All neural models are implemented in PyTorch~1.9.1 and trained with CUDA~11.1 and cuDNN~8.0.5.

Table~\ref{tab:main-exp-settings} summarizes the main settings for data construction and HAR model training.
The main HAR results are averaged over three random seeds (45, 46, and 47) using the same LOSO splits.
Some ablation studies use different random-seed sets from the main experiments; their absolute values should therefore be compared within the corresponding table or figure.
Within each table or figure, all compared configurations use the same random seeds and evaluation protocol.

\begin{table}[t]
\centering
\caption{Main settings for virtual IMU construction and HAR training.}
\label{tab:main-exp-settings}
\scriptsize
\setlength{\tabcolsep}{2pt}
\renewcommand{\arraystretch}{1.12}
\begin{tabular}{p{0.34\linewidth}p{0.57\linewidth}}
\hline
Aspect & Setting \\
\hline
Data preprocessing & 2-s windows; 1-s stride; 20 Hz \\
Anchor selection & class-wise diversity and scarcity anchors from real training windows \\
\hline
Prompt conditioning & activity label; tempo, intensity, periodicity \\
Virtual generation & T2M-GPT followed by IMUSim \\
Candidate pool & 20 candidates per anchor \\
\hline
Candidate selection & lowest-cost $K_{\mathrm{sel}}$ candidates based on embedding distance and label consistency \\
Training integration & anchor-level risk and within-anchor candidate rank \\
Evaluation & LOSO; each held-out subject is used only for evaluation \\
\hline
\end{tabular}
\end{table}

\subsubsection{Training Configurations}

For the three HAR benchmarks, we evaluate the following six training configurations under the LOSO protocol.
Within each benchmark and downstream model, all configurations use identical data splits, preprocessing, model settings, and evaluation procedures; they differ only in how additional training data are constructed and integrated.

\begin{enumerate}
    \item \textbf{Real-only.} The model is trained only on the real IMU training data, without any virtual data or augmentation. This serves as the primary supervised baseline.

    \item \textbf{Traditional Augmentation~\cite{leng2025scaling}.} The model is trained on real IMU data with standard sensor-level augmentation, including rotation, Gaussian noise, and additive sensor bias.

    \item \textbf{TimeGAN~\cite{yoon2019time}.} The training set is augmented with synthetic IMU sequences generated using the official open-source implementation of TimeGAN, trained separately on the real data in each LOSO training fold.

    \item \textbf{Diffusion-TS~\cite{yuan2024diffusionts}.} For the Diffusion-TS baseline, we use the official Diffusion-TS implementation~\cite{yuan2024diffusionts} with a class-wise unconditional adaptation, training a separate generator for each activity class within each LOSO fold using only the corresponding real training windows.

    \item \textbf{IMUGPT~\cite{leng2024imugpt}.} The model is trained on real IMU data augmented with IMUGPT-style sequences generated from activity-level prompts.

    \item \textbf{Ours.} The full method trains on real data and anchor-conditioned virtual IMU candidates, with their selection costs determining candidate retention and training weights.
\end{enumerate}

For a fair comparison, all configurations using additional data are limited to 150 sequences per activity class before window segmentation.

\subsection{Main Results}

Table~\ref{tab:main-results} summarizes the main HAR results on USC-HAD, PAMAP2, and RealWorld using DeepConvLSTM, DeepConvLSTM-Attention, and MLP-HAR.
Values are reported as percentages in the form mean~$\pm$~standard deviation.
F1 denotes Macro-F1, and $\Delta$F1 denotes the absolute Macro-F1 improvement over the Real-only baseline.
The reported results use 20\% of the labeled real data on USC-HAD and PAMAP2 and 10\% on RealWorld.

\begin{table*}[t]
\centering
\caption{Main results across three downstream HAR models.}
\label{tab:main-results}
\begin{threeparttable}
\scriptsize
\setlength{\tabcolsep}{1.4pt}
\renewcommand{\arraystretch}{1.05}
\begin{tabular}{@{}llccc@{\hskip 7pt}ccc@{\hskip 7pt}ccc@{}}
\toprule
\multirow{2}{*}{\textbf{Dataset}} & \multirow{2}{*}{\textbf{Method}}
& \multicolumn{3}{c}{\textbf{DeepConvLSTM}~\cite{ordonez2016deep}}
& \multicolumn{3}{c}{\textbf{DeepConvLSTM-Attention}~\cite{murahari2018attention}}
& \multicolumn{3}{c}{\textbf{MLP-HAR}~\cite{zhou2024mlp}} \\
\cmidrule(lr){3-5}\cmidrule(lr){6-8}\cmidrule(l){9-11}
& & \textbf{Acc. $\uparrow$} & \textbf{F1 $\uparrow$} & \textbf{$\Delta$F1}
& \textbf{Acc. $\uparrow$} & \textbf{F1 $\uparrow$} & \textbf{$\Delta$F1}
& \textbf{Acc. $\uparrow$} & \textbf{F1 $\uparrow$} & \textbf{$\Delta$F1} \\
\midrule
\multirow{6}{*}{USC-HAD}
& Real-only & 75.44 $\pm$ 3.05 & 74.18 $\pm$ 1.99 & -- & 79.21 $\pm$ 1.17 & 78.28 $\pm$ 1.63 & -- & 58.20 $\pm$ 0.63 & 54.14 $\pm$ 0.97 & -- \\
& Trad. Aug.~\cite{leng2025scaling} & 80.23 $\pm$ 0.44 & 78.94 $\pm$ 0.87 & +4.76 & 81.08 $\pm$ 0.59 & 80.25 $\pm$ 0.40 & +1.97 & 62.22 $\pm$ 1.27 & 58.26 $\pm$ 1.22 & +4.12 \\
& TimeGAN~\cite{yoon2019time} & 77.47 $\pm$ 1.15 & 76.20 $\pm$ 1.38 & +2.02 & 79.31 $\pm$ 1.43 & 78.05 $\pm$ 1.07 & -0.23 & 60.05 $\pm$ 1.30 & 55.41 $\pm$ 1.48 & +1.27 \\
& Diffusion-TS~\cite{yuan2024diffusionts} & 80.13 $\pm$ 2.47 & 78.77 $\pm$ 2.29 & +4.59 & 80.33 $\pm$ 1.05 & 79.22 $\pm$ 0.87 & +0.94 & 63.18 $\pm$ 1.33 & 59.99 $\pm$ 1.51 & +5.84 \\
& IMUGPT~\cite{leng2024imugpt} & 75.41 $\pm$ 1.33 & 73.02 $\pm$ 0.98 & -1.16 & 75.06 $\pm$ 0.36 & 73.22 $\pm$ 0.26 & -5.06 & 61.43 $\pm$ 0.56 & 58.59 $\pm$ 0.64 & +4.45 \\
& \textbf{Ours} & \textbf{82.93 $\pm$ 0.66} & \textbf{80.99 $\pm$ 0.28} & \textbf{+6.80} & \textbf{81.97 $\pm$ 0.90} & \textbf{80.43 $\pm$ 1.12} & \textbf{+2.15} & \textbf{67.04 $\pm$ 1.81} & \textbf{65.38 $\pm$ 1.34} & \textbf{+11.24} \\
\midrule
\multirow{6}{*}{PAMAP2}
& Real-only & 70.97 $\pm$ 0.11 & 62.95 $\pm$ 1.07 & -- & 75.36 $\pm$ 2.55 & 68.12 $\pm$ 1.88 & -- & 68.58 $\pm$ 7.56 & 53.80 $\pm$ 6.53 & -- \\
& Trad. Aug. & 79.58 $\pm$ 1.85 & 72.17 $\pm$ 3.41 & +9.22 & 80.50 $\pm$ 1.18 & 74.47 $\pm$ 2.44 & +6.35 & 70.06 $\pm$ 7.76 & 57.20 $\pm$ 6.74 & +3.40 \\
& TimeGAN & 77.64 $\pm$ 1.83 & 68.25 $\pm$ 1.73 & +5.30 & 80.27 $\pm$ 0.86 & 72.93 $\pm$ 4.07 & +4.81 & 67.71 $\pm$ 6.99 & 54.50 $\pm$ 5.93 & +0.70 \\
& Diffusion-TS & 80.78 $\pm$ 0.49 & 68.89 $\pm$ 1.64 & +5.94 & 80.59 $\pm$ 0.98 & 68.81 $\pm$ 3.28 & +0.69 & 68.50 $\pm$ 7.50 & 54.85 $\pm$ 6.96 & +1.04 \\
& IMUGPT & 73.89 $\pm$ 0.50 & 65.83 $\pm$ 3.06 & +2.88 & 73.98 $\pm$ 1.16 & 63.68 $\pm$ 2.25 & -4.44 & 71.18 $\pm$ 7.37 & 57.92 $\pm$ 6.65 & +4.12 \\
& \textbf{Ours} & \textbf{80.98 $\pm$ 0.33} & \textbf{77.35 $\pm$ 0.51} & \textbf{+14.40} & \textbf{81.92 $\pm$ 0.72} & \textbf{77.25 $\pm$ 1.72} & \textbf{+9.13} & \textbf{72.51 $\pm$ 7.78} & \textbf{59.08 $\pm$ 8.72} & \textbf{+5.28} \\
\midrule
\multirow{6}{*}{RealWorld}
& Real-only & 65.75 $\pm$ 1.43 & 66.76 $\pm$ 0.95 & -- & 68.79 $\pm$ 1.32 & 68.84 $\pm$ 1.63 & -- & 62.07 $\pm$ 1.34 & 60.14 $\pm$ 3.12 & -- \\
& Trad. Aug. & 68.33 $\pm$ 2.12 & 69.18 $\pm$ 2.03 & +2.42 & 71.27 $\pm$ 1.27 & 71.85 $\pm$ 0.51 & +3.01 & 62.89 $\pm$ 1.25 & 63.10 $\pm$ 1.48 & +2.96 \\
& TimeGAN & 68.58 $\pm$ 1.42 & 68.88 $\pm$ 1.07 & +2.12 & 69.99 $\pm$ 1.23 & 70.82 $\pm$ 1.14 & +1.98 & 62.21 $\pm$ 1.24 & 61.38 $\pm$ 1.79 & +1.24 \\
& Diffusion-TS & 68.68 $\pm$ 0.96 & 68.34 $\pm$ 0.81 & +1.58 & 70.91 $\pm$ 0.67 & 70.96 $\pm$ 0.89 & +2.12 & 62.43 $\pm$ 0.79 & 62.40 $\pm$ 1.14 & +2.26 \\
& IMUGPT & 66.29 $\pm$ 1.23 & 65.50 $\pm$ 1.31 & -1.26 & 69.05 $\pm$ 1.35 & 68.26 $\pm$ 1.40 & -0.58 & 60.44 $\pm$ 0.86 & 59.45 $\pm$ 0.79 & -0.69 \\
& \textbf{Ours} & \textbf{71.65 $\pm$ 1.72} & \textbf{71.68 $\pm$ 1.45} & \textbf{+4.92} & \textbf{72.47 $\pm$ 1.18} & \textbf{72.61 $\pm$ 1.77} & \textbf{+3.77} & \textbf{64.33 $\pm$ 1.28} & \textbf{64.00 $\pm$ 1.06} & \textbf{+3.86} \\
\bottomrule
\end{tabular}
\end{threeparttable}
\end{table*}

For every combination of dataset and model, \textbf{Ours} achieves the highest mean Macro-F1 and accuracy.
Compared with Real-only, it improves Macro-F1 by 2.15 to 11.24 percentage points on USC-HAD, 5.28 to 14.40 points on PAMAP2, and 3.77 to 4.92 points on RealWorld.
It also outperforms the strongest non-Ours baseline by 0.18 to 5.39 points on USC-HAD, 1.16 to 5.18 points on PAMAP2, and 0.76 to 2.50 points on RealWorld.
These results show that the method remains effective across convolutional-recurrent, attention-based, and feed-forward HAR models, although the gains vary across datasets and downstream models.

On PAMAP2, the gains over the strongest non-Ours baseline are 5.18, 2.78, and 1.16 points with DeepConvLSTM, DeepConvLSTM-Attention, and MLP-HAR, respectively.
With DeepConvLSTM, Ours improves Macro-F1 from 62.95\% to 77.35\% and exceeds the strongest competing baseline by 5.18 points.
On USC-HAD, Ours shows a large improvement over the strongest non-Ours baseline with DeepConvLSTM and MLP-HAR.
However, it is only 0.18 points higher than Traditional Augmentation with DeepConvLSTM-Attention.
On RealWorld, the gains over the strongest non-Ours baselines are smaller but remain positive across all three downstream models.

Traditional Augmentation is the strongest non-Ours Macro-F1 baseline in seven of the nine dataset--model pairs.
Diffusion-TS improves over Real-only in all nine settings, and TimeGAN does so in eight, whereas IMUGPT improves Macro-F1 in only three of the nine settings.
Under the common sequence-level data budget, none of the evaluated synthetic-generation baselines consistently outperforms Traditional Augmentation.

\subsection{Ablation Study}

\begin{figure}[!t]
\centering
\includegraphics[width=\columnwidth]{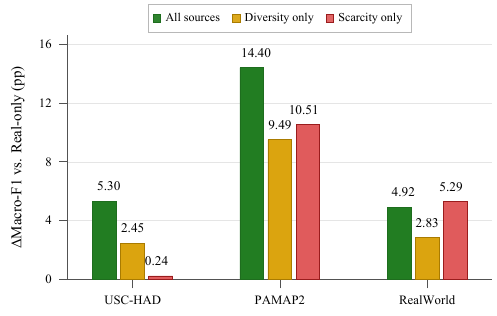}
\caption{Effect of anchor-source composition. Bars show the Macro-F1 gain over the corresponding Real-only baseline. Combining diversity and scarcity anchors performs best on USC-HAD and PAMAP2, whereas scarcity-only is slightly higher on RealWorld.}
\label{fig:anchor-source-composition}
\end{figure}

\paragraph{Coverage-aware vs. random anchor selection.}
Table~\ref{tab:anchor-targeting-ablation} compares the full method with a random class-balanced anchor variant, while keeping the remaining pipeline unchanged, to isolate the effect of coverage-aware anchor selection.
The random-anchor variant still outperforms Real-only on all three datasets, showing that virtual candidates can benefit HAR training even when their generation anchors are selected randomly.
However, the full method consistently outperforms random anchors, with additional Macro-F1 gains of 0.76, 3.41, and 0.70 percentage points on USC-HAD, PAMAP2, and RealWorld, respectively.
The larger gain on PAMAP2 suggests that coverage-aware anchor selection is more beneficial on this dataset.
Since both variants use the same virtual-data budget, the improvement over random selection shows that the choice of generation anchors matters beyond the number of generated samples.

\begin{table}[t]
\centering
\caption{Coverage-aware versus random anchor selection.}
\label{tab:anchor-targeting-ablation}
\begin{threeparttable}
\scriptsize
\setlength{\tabcolsep}{2.0pt}
\renewcommand{\arraystretch}{1.05}
\begin{tabular}{@{}lcccc@{}}
\toprule
\textbf{Dataset} & \textbf{Real-only} & \textbf{Random} & \textbf{Full Ours} & \textbf{$\Delta$F1} \\
\midrule
USC-HAD & 74.32 $\pm$ 1.93 & 80.18 $\pm$ 0.81 & \textbf{80.94 $\pm$ 0.74} & +0.76 \\
PAMAP2 & 62.95 $\pm$ 1.07 & 73.94 $\pm$ 0.50 & \textbf{77.35 $\pm$ 0.51} & +3.41 \\
RealWorld & 66.97 $\pm$ 1.12 & 70.98 $\pm$ 1.74 & \textbf{71.68 $\pm$ 1.45} & +0.70 \\
\bottomrule
\end{tabular}
\begin{tablenotes}[flushleft]
\footnotesize
\item $\Delta$F1 is the absolute Macro-F1 improvement of Full Ours over random class-balanced anchors.
\end{tablenotes}
\end{threeparttable}
\end{table}

\paragraph{Anchor-source composition.}
Fig.~\ref{fig:anchor-source-composition} compares the effects of diversity-only, scarcity-only, and combined anchor-source configurations across the three datasets.
On USC-HAD, the diversity-only and scarcity-only variants improve Macro-F1 over Real-only by 2.45 and 0.24 percentage points, respectively, while the combined variant increases the gain to 5.30 points.
Diversity anchors are the stronger single source in this setting.
In particular, the 5.30-point gain of the combined variant is larger than the sum of the gains from the two single-source variants (2.69 points).
This result suggests that, although scarcity anchors contribute little when used alone in this setting, they may provide additional local coverage or training stability when combined with diversity anchors.
On PAMAP2, the diversity-only and scarcity-only variants improve over Real-only by 9.49 and 10.51 points, respectively, while their combination achieves the largest gain of 14.40 points.
On RealWorld, the scarcity-only variant achieves the largest gain of 5.29 points and exceeds the combined variant by 0.37 points.
This result shows that adding another anchor source does not always lead to better performance.

Overall, the contribution of each anchor source is dataset dependent. Diversity is the stronger single source on USC-HAD, both sources are effective on PAMAP2, and scarcity is the strongest source on RealWorld.
Given that RealWorld was collected under less controlled daily-life conditions, the higher F1 of scarcity-only suggests that targeting sparsely represented regions is particularly useful on this dataset.
These results indicate that both anchor sources are useful in the coverage-aware framework, but their contributions differ across datasets.

\paragraph{Cost-based vs. order-based candidate selection.}

\begin{figure}[!t]
\centering
\includegraphics[width=\columnwidth]
{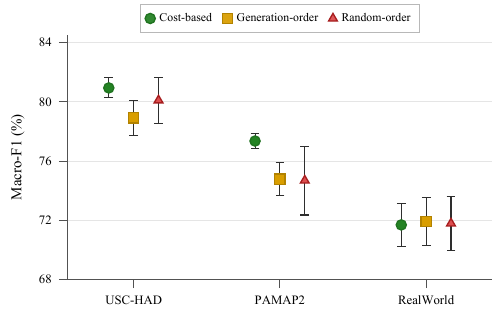}
\caption{Effect of candidate ordering across the three datasets.}
\label{fig:candidate-ordering}
\end{figure}

To test whether candidate-level scoring improves the usefulness of virtual samples for HAR training, we compare three selection strategies on the same candidate pool generated for each anchor.
All three strategies operate on the same pool of up to $M$ generated candidates.
Cost-based selection ranks candidates by the selection cost defined from anchor proximity and target-label consistency and uses the $K_{\mathrm{sel}}$ candidates with the lowest costs.
Generation-order does not use the cost and selects the first $K_{\mathrm{sel}}$ valid candidates in their original generation order.
Random-order shuffles the valid candidates using a fixed random seed and uses the first $K_{\mathrm{sel}}$ candidates.
Within each dataset, the three strategies use identical anchors, generated candidate pools, candidate budgets, candidate post-processing, reliability weights, and downstream HAR settings; only the candidate ordering and resulting selection differ.

On USC-HAD, Cost-based achieves a Macro-F1 of 80.94\%, exceeding Generation-order and Random-order by 2.02 and 0.84 percentage points.
Both order-based variants still outperform Real-only, showing that virtual candidates can benefit HAR training even without cost-based ordering. On USC-HAD, ordering candidates by selection cost provides a further gain.
On PAMAP2, Cost-based achieves the highest mean Macro-F1 of 77.35\%, exceeding the order-based baseline by 2.58 percentage points.
On RealWorld, Generation-order achieves 71.92\%, while Random-order achieves 71.78\%.
Both perform slightly better than Cost-based at 71.68\%.
Cost-based also has the lowest standard deviation on all three datasets: 0.68 on USC-HAD, 0.51 on PAMAP2, and 1.45 on RealWorld.
Overall, Cost-based selection improves mean Macro-F1 on USC-HAD and PAMAP2 and shows the lowest observed variability on all three datasets.
These results show that anchor proximity and target-label consistency can provide useful information for candidate selection, but their effectiveness varies across datasets.

\paragraph{Rank-aware vs. equal-rank weighting.}

\begin{table}[t]
\centering
\caption{Comparison of rank-aware and equal-rank weighting. Values are Macro-F1 (\%); $\Delta$ is computed as Equal-rank minus Full Ours.}
\label{tab:rank-weighting-ablation}
\small
\setlength{\tabcolsep}{3.0pt}
\renewcommand{\arraystretch}{1.05}
\begin{tabular}{lccc}
\toprule
\textbf{Dataset} & \textbf{Full Ours} & \textbf{Equal-rank} & \textbf{$\Delta$} \\
\midrule
USC-HAD
& \textbf{80.94 $\pm$ 0.68}
& 78.71 $\pm$ 1.78
& -2.23 \\
PAMAP2
& \textbf{77.35 $\pm$ 0.51}
& 73.59 $\pm$ 0.18
& -3.76 \\
RealWorld
& \textbf{71.68 $\pm$ 1.45}
& 70.13 $\pm$ 1.95
& -1.55 \\
\bottomrule
\end{tabular}
\end{table}

Table~\ref{tab:rank-weighting-ablation} evaluates whether candidate rank remains informative after candidate selection.
Full Ours and Equal-rank use the same selected candidates and differ only in their within-anchor rank weighting.
Full Ours assigns decreasing rank multipliers, whereas Equal-rank assigns the same multiplier to all selected candidates from the same anchor.
This comparison tests whether selected candidates should receive equal or rank-dependent training weights.

Compared with the Full Ours reference, Equal-rank weighting is lower by 2.23, 3.76, and 1.55 percentage points in mean Macro-F1 on USC-HAD, PAMAP2, and RealWorld.
The lower results under Equal-rank weighting indicate that the retained candidates should not receive identical weights.
Higher-ranked candidates should receive larger weights than lower-ranked candidates.

\subsection{Data Efficiency Analysis}

\paragraph{Label efficiency.}
\begin{figure*}[!t]
\centering
\includegraphics[width=\textwidth]{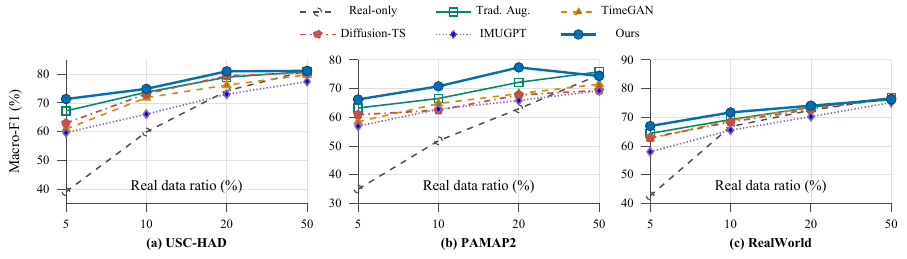}
\caption{Macro-F1 under different labeled real-data ratios on (a) USC-HAD, (b) PAMAP2, and (c) RealWorld.}
\label{fig:data-efficiency}
\end{figure*}

Fig.~\ref{fig:data-efficiency} compares Macro-F1 on USC-HAD, PAMAP2, and RealWorld at labeled real-data ratios of 5\%, 10\%, 20\%, and 50\%.
For each method, we vary only the labeled real-data ratio in each LOSO training fold while keeping its augmentation or virtual-data pipeline and downstream training settings unchanged.
At the 5\% ratio, Ours improves over Real-only by 32.28 points on USC-HAD, 31.24 points on PAMAP2, and 24.32 points on RealWorld.
The gains decrease to 15.03, 18.95, and 4.92 points at 10\%, and to 6.81, 14.40, and 1.57 points at 20\%.
At 50\%, the differences between Ours and Real-only are $-0.06$, $-0.39$, and $-0.38$ points, leaving their mean performance nearly identical.
Other methods show a similar trend.
At 5\%, Diffusion-TS improves over Real-only by 23.87 points on USC-HAD, 25.92 points on PAMAP2, and 20.25 points on RealWorld.
At 50\%, the corresponding differences are $-0.63$, $-5.40$, and 0.02 points.
The largest gains from augmentation and virtual data occur at low labeled real-data ratios.
Ours achieves the highest mean Macro-F1 on all three datasets at 5\%, 10\%, and 20\%, but its advantage over Real-only narrows as more labeled real data are used and is almost absent at 50\%.
Other methods show the same trend, and some settings exhibit slight negative transfer at higher labeled ratios.
Virtual IMU data are most useful when labeled real data are limited; their benefit becomes small or negative as the real training set grows.

\paragraph{Virtual-data scaling.}

\begin{figure}[!t]
\centering
\includegraphics[width=\columnwidth]
{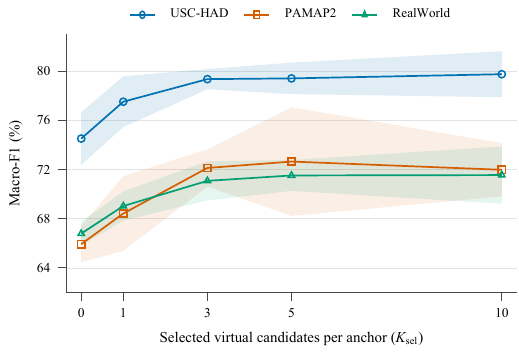}
\caption{Macro-F1 with different numbers of selected virtual candidates per anchor on USC-HAD, PAMAP2, and RealWorld. Error bars indicate the standard deviation across random seeds.}
\label{fig:virtual-budget-selection}
\end{figure}

Fig.~\ref{fig:virtual-budget-selection} compares Macro-F1 with different numbers of selected virtual candidates per anchor.
We fix the labeled real-data subset and vary only the maximum number of selected candidates per anchor, $K_{\mathrm{sel}} \in \{0, 1, 3, 5, 10\}$, where $K_{\mathrm{sel}}=0$ corresponds to Real-only.
USC-HAD and PAMAP2 use 20\% labeled real data, whereas RealWorld uses 10\%.
All settings reuse the same candidate pool and ranking, and candidates shared across settings use the same training weights.
With only the top-ranked candidate retained for each anchor, Macro-F1 improves over Real-only by 3.00 points on USC-HAD, 2.49 points on PAMAP2, and 2.24 points on RealWorld.
At $K_{\mathrm{sel}}=3$, the gains increase to 4.84, 6.20, and 4.29 points.
Further increases in $K_{\mathrm{sel}}$ bring only small changes in performance.
On USC-HAD, Macro-F1 increases from 79.36\% $\pm$ 0.82\% at $K_{\mathrm{sel}}=3$ to 79.76\% $\pm$ 1.87\% at $K_{\mathrm{sel}}=10$.
On PAMAP2, Macro-F1 reaches 72.65\% $\pm$ 4.42\% at $K_{\mathrm{sel}}=5$ and then decreases to 71.98\% $\pm$ 2.17\% at $K_{\mathrm{sel}}=10$.
On RealWorld, Macro-F1 reaches 71.52\% $\pm$ 1.26\% at $K_{\mathrm{sel}}=5$ and increases by only 0.03 points when $K_{\mathrm{sel}}$ is increased to 10.

Most of the improvement comes from the first few high-ranked candidates.
At $K_{\mathrm{sel}}=3$, USC-HAD and RealWorld already obtain about 92\% and 90\% of the total improvement observed at $K_{\mathrm{sel}}=10$, while PAMAP2 performs slightly better than at $K_{\mathrm{sel}}=10$.
Thus, more virtual candidates do not necessarily lead to better performance.
The results show that most of the gains come from a small number of top-ranked candidates, while adding lower-ranked candidates provides little further benefit and can even slightly degrade performance on some datasets.

\subsection{Feature Visualization}

\begin{figure*}[t]
    \centering
    \includegraphics[width=\textwidth]{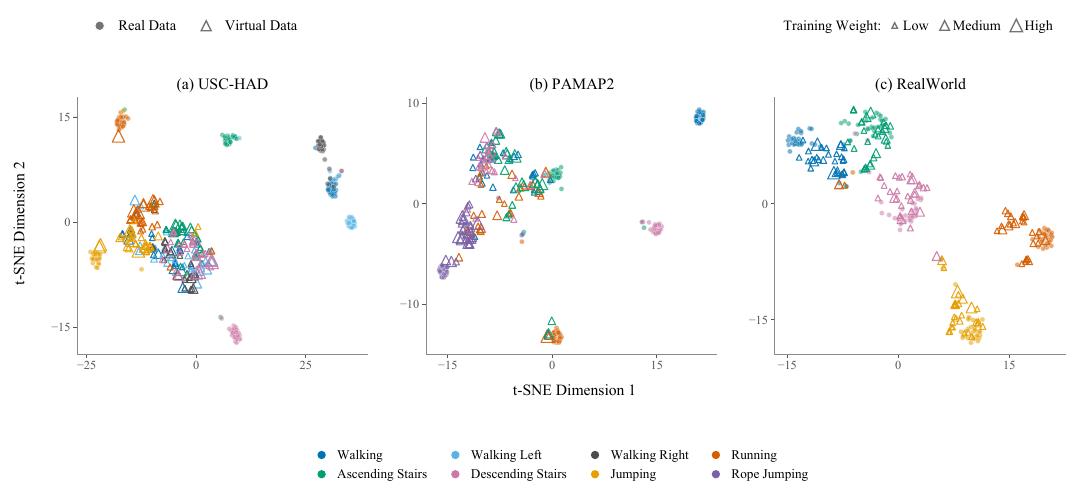}
    \caption{t-SNE visualization of low-label real training data and final selected virtual IMU samples on (a) USC-HAD, (b) PAMAP2, and (c) RealWorld. Colors denote activity classes, circles and triangles indicate real and virtual samples, respectively, and triangle size represents the effective training weight of each virtual sample.}
    \label{fig:virtual-imu-tsne}
\end{figure*}

Fig.~\ref{fig:virtual-imu-tsne} visualizes the learned representations of real training samples and selected virtual IMU samples on USC-HAD, PAMAP2, and RealWorld.
For each dataset, we use one fixed LOSO outer fold and jointly project the learned representations of its real training and virtual samples into a two-dimensional t-SNE space; samples from the held-out subject are excluded.
Colors denote activity classes, circles and triangles denote real and virtual samples, respectively, and triangle size indicates the effective training weight.

The three datasets show different relationships between real and virtual samples.
On USC-HAD, real and virtual samples show noticeable separation, and virtual representations from different activities overlap substantially.
On PAMAP2, the virtual samples show clearer class structure, although some remain separated from their corresponding real clusters.
On RealWorld, real and virtual samples from the same activity often occupy nearby regions, with weaker separation by data source.
The selected virtual samples show activity-related structure in the learned representation space.
Their correspondence with real samples varies across datasets, consistent with the main results in the full LOSO evaluation.
The t-SNE plots show how the selected virtual samples relate to real training samples, while the
full LOSO evaluation measures their effect on HAR performance.



\subsection{Exploratory Extension to Medical Wearable IMU}
Beyond the standard HAR benchmarks, we conduct an exploratory evaluation on PADS~\cite{varghese2024machine,PhysioNet-parkinsons-disease-smartwatch-1.0.0}, a wearable IMU dataset for Parkinson's disease (PD) versus healthy-control (HC) classification.
The evaluation uses both-wrist recordings from the StretchHold and HoldWeight tasks, with 20\% labeled real data.
We use a subject-level five-fold split and an RBF-SVM classifier; hyperparameters are selected by inner cross-validation using only the training subjects in each outer fold.

Table~\ref{tab:pads-medical-extension} reports the PADS results.
Compared with Real-only, Ours improves Balanced Accuracy from 63.64\% to 80.52\% and Macro-F1 from 65.02\% to 78.94\%, corresponding to gains of 16.88 and 13.92 percentage points, respectively.
Diffusion-TS reaches 77.27\% $\pm$ 4.31\% Balanced Accuracy and 77.22\% $\pm$ 4.53\% Macro-F1, improving over Real-only by 13.63 and 12.19 percentage points, respectively.
It outperforms Traditional Augmentation and TimeGAN on both metrics but remains below IMUGPT and Ours.
Ours obtains the highest mean results, exceeding IMUGPT by 0.32 points in Balanced Accuracy and 0.66 points in Macro-F1.
These results indicate that the proposed virtual IMU pipeline can extend to a low-label medical wearable classification task.

\begin{table}[t]
\centering
\begin{threeparttable}
\caption{PD-vs-HC classification results on PADS.}
\label{tab:pads-medical-extension}
\footnotesize
\setlength{\tabcolsep}{2.5pt}
\begin{tabular}{@{}lccc@{}}
\toprule
Method & Bal. Acc. $\uparrow$ & Macro-F1 $\uparrow$ & $\Delta$Macro-F1 \\
\midrule
Real-only
& $63.64 \pm 8.90$
& $65.02 \pm 11.30$
& -- \\
Trad. Aug.
& $73.82 \pm 10.47$
& $74.94 \pm 10.10$
& $+9.92$ \\
TimeGAN
& $76.82 \pm 9.84$
& $73.29 \pm 10.36$
& $+8.27$ \\
Diffusion-TS
& $77.27 \pm 4.31$
& $77.22 \pm 4.53$
& $+12.19$ \\
IMUGPT
& $80.20 \pm 8.76$
& $78.28 \pm 7.74$
& $+13.26$ \\
Ours
& $\mathbf{80.52 \pm 9.08}$
& $\mathbf{78.94 \pm 8.64}$
& $\mathbf{+13.92}$ \\
\bottomrule
\end{tabular}
\begin{tablenotes}[flushleft]
\footnotesize
\item Results use 20\% labeled real data. Values are percentages reported as mean $\pm$ standard deviation across five subject-level outer folds. $\Delta$F1 is computed from unrounded Macro-F1 values relative to Real-only.
\end{tablenotes}
\end{threeparttable}

\end{table}

\section{Limitations and Future Work}

\subsection{Limitations}

\paragraph{Dependence on the upstream motion generator.}
Our current implementation uses a fixed text-to-motion model to produce motions for virtual IMU synthesis.
The scope and quality of the resulting candidate pool depend on the motion patterns represented by this generator.
Motion patterns that are weakly represented in its training data may be generated less reliably.

\paragraph{Simplified sensor simulation.}
Our current pipeline uses IMUSim to generate ideal accelerometer and gyroscope signals at the skeleton joint nearest to each target sensor location.
It does not explicitly model changes in sensor placement and orientation during wear or device-specific noise and bias.
These factors can change the measured IMU signals even when the underlying body motion is the same.

\paragraph{Limited real-world and clinical evaluation.}
All experiments in this study are conducted offline using existing datasets.
We have not evaluated the method in long-term deployments or under changes in real-world data collection conditions.
The PADS experiment provides an exploratory evaluation of PD-versus-HC classification rather than a clinical validation.

\subsection{Future Work}

\paragraph{Sensor-aware virtual IMU generation.}
Future work could extend virtual IMU generation beyond a fixed motion-to-IMU conversion.
The generation process could model sensor placement, orientation, and device-specific noise and bias.
The goal is to make virtual IMU signals reflect changes in sensor configuration, rather than simply generating more candidates.

\paragraph{Self-evolving agent for virtual IMU generation.}
A self-evolving agent could manage virtual IMU generation, evaluation, and refinement.
The agent could record effective generation strategies and common failure patterns across activities.
It could then update prompts, sampling rules, and candidate-selection criteria based on this experience.
The agent could reuse and refine these strategies across activities and datasets.

\paragraph{Extending virtual IMU generation to clinical events.}
Future work could apply virtual IMU generation to transient clinical events such as freezing of gait.
For freezing of gait, generated sequences should cover the transition into and out of each episode, rather than only the event segment.
The generation process could also include hard negatives such as slow walking, hesitation, and brief stops.
This would add training examples near event boundaries instead of only increasing the number of positive events.
Models trained with these data could be evaluated on event localization, duration estimation, and false-alarm rate.

\section{Conclusion}
In this paper, we proposed a coverage-aware virtual IMU augmentation framework for HAR under limited real training data.
The framework selects diversity and scarcity anchors in a learned sensor embedding space, converts their dynamics attributes into prompts for virtual IMU generation, and assesses generated candidates using anchor proximity and label consistency.
The selected candidates are then integrated into downstream HAR training with weights determined by their estimated reliability.
Experiments on three public HAR benchmarks show consistent improvements over competitive baselines under limited labeled-data settings, while ablation studies examine where virtual candidates are generated, which candidates are retained, and how strongly they influence training.
In the future, we will further study how to improve the reliability of virtual IMU generation under more diverse real-world sensing conditions.

\bibliographystyle{IEEEtran}
\bibliography{references}

\end{document}